# Connectionism, Complexity, and Living Systems: a comparison of Artificial and Biological Neural Networks

Krishna Katyal[1,3], Jesse Parent[2], and Bradly Alicea[2,3]
[1] Shri Mata Vaishno Devi University, Jammu and Kashmir  INDIA
[2] Orthogonal Research and Education Lab, Champaign, IL  USA
[3] OpenWorm Foundation, Boston, MA  USA

**Abstract**
While Artificial Neural Networks (ANNs) have yielded impressive results in the realm of simulated intelligent behavior, it is important to remember that they are but sparse approximations of Biological Neural Networks (BNNs). We go beyond comparison of ANNs and BNNs to introduce principles from BNNs that might guide the further development of ANNs as embodied neural models. These principles include representational complexity, complex network structure/energetics, and robust function. We then consider these principles in ways that might be implemented in the future development of ANNs. In conclusion, we consider the utility of this comparison, particularly in terms of building more robust and dynamic ANNs. This even includes constructing a morphology and sensory apparatus to create an embodied ANN, which when complemented with the organizational and functional advantages of BNNs unlocks the adaptive potential of lifelike networks.

**Introduction**

How can Artificial Neural Networks (ANNs) emulate the "lifelike" nature of Biological Neural Networks (BNNs)? In recent years, flavors of ANN such as Deep Neural Networks (DNNs), Generative Adversarial Networks (GANs), and Convolutional Neural Networks (CNNs) have been trained to produce generative and ephemeral outputs we consider to be lifelike. For example, GANs have enabled procedural generation (Risi and Togelius, 2020), which allows for the creation of art and other creative content. The one-shot language learning model GPT-3 (Brown, et al. 2020) is based on a transformer neural network and exhibits impressive performance. Based on these advances, one might think that a sparse representation of the brain is sufficient to approximate intelligent behavior.

Yet there are limits to the realism of the outputs of such models. In cases where human users interact with procedurally-generated virtual characters, the human response resembles the uncanny valley effect (Tinwell, et al. 2013). Similarly, GPT-3 can exhibit strange and often dangerously incorrect assumptions about the world in which it is situated (Marcus and Davis, 2020). This suggests that future improvements should be in terms of fundamental shortcomings of the ANN model, and potential solutions including various forms of biological inspiration. Our goal here is to deconstruct ANNs in terms of their parallels (or lack thereof) with BNNs. In doing so, we wish to better understand properties that might make ANNs more like living systems. These properties may not necessarily improve the performance of ANNs, but they might afford higher order properties such as representational complexity, the relationship between energetics, information processing, and complex network structure, and an increase in robust function.

## Comparison of ANNs and BNNs

### Sparse Behavioral Output and Internal Representations of ANNs

One fundamental problem with mapping ANNs to intelligent behavior is that neural networks were never meant to be behavioral engines, but rather explicit models of the internal world (Rumelhart and McClelland, 1986). As models of the internal milieu, however, they are emergentist structures which drive complex output behaviors nonetheless (Marco and Alberto, 2018; Wu, et al. 2020). Compare this to the reductionist nature of a Central Pattern Generator (CPG): a phenomenological internal model that produces easily interpretable behavioral output (Selverston, 1980).

The standard representation of an ANN exhibits three major differences as compared to BNNs. The first is that ANNs exploit computational parsimony over the complexity of neural biology. ANNs typically consist only of units (generic cell types), connections (an amalgam of axons and synapses), connection strengths (analogous to chemical connectivity), and activation functions (a stand-in for action potentials). By contrast, BNNs typically include biochemically complex neurons with support cells and dense connections across many other neurons. Critically, BNNs are hybrid digital-analog systems, which differs from the typically digital nature of ANNs. These properties point to a second difference, which is a homogenization of neural communication in ANNs versus BNNs. In ANNs, cells have no functional difference, and the structure of these networks is uniform. The third major difference involves energetic capacity and, perhaps more importantly, energetic constraints. This leads to the fourth difference, which is robust function in the face of ambiguous inputs or a lack of context.

### Representational Complexity

Drawing upon the notion of computational parsimony, ANNs are famous for their minimal representation (Rogers and McClelland, 2014; Hassabis, et al. 2017). This is particularly true with respect to BNNs (Figure 1). While ANNs provide a means for approximating the parallelism and combinatorial connectivity of the brain, they do not approximate the great complexity of neural tissues, or even the complexity of circuits such as those found in the retina or cerebellum. From a systems standpoint, the retina is but one example of a circuit that exhibits something we might refer to as neural spaghettification (Alicea, 2011). While the cells are arranged in layers, signals travel run a circuitous route which results from evolutionary processes (Gregory, 2008). Another example are the convolutions that define the surface of the Mammalian neocortex. In this case, we not only have a great complexity of cell types, electrical and chemical connections, and intercellular matrix, but also folds that exhibit a fractal dimensionality (Im, et al. 2006). This has implications for the BNN topology, which we will discuss later.

This does not mean that ANNs are inferior to BNNs. In fact, ANNs and BNNs in the form of a Zebrafish connectome can yield convergent representations of environmental phenomena such as temperature (Haesemeyer, et al. 2019). One critical difference between ANNs and BNNs is the role of static training and biological plasticity. ANNs are defined by their thirst for data, namely labelled data. By contrast, BNNs can not only learn without prior knowledge, but also emerge from developmental precursors. Therefore, one major factor in allowing BNNs to deal with naive learning is the presence of an innate component. Indeed, ANNs do not have innate or biologically systemic components such as genomes or extended developmental and growth processes (Zador, 2019). While neuroevolution approaches (Angeline, et al. 1994; Stanley and Miikkulainen, 2002) apply evolutionary computing techniques to ANNs, they can only roughly approximate the complex biological substrate. The neurons of BNNs are reliant upon not only gene expression, but neurotransmitter dynamics and protein synthesis as well. Depending on the species, BNNs produce both reflexive, non-learned behaviors and complex acquired behaviors.

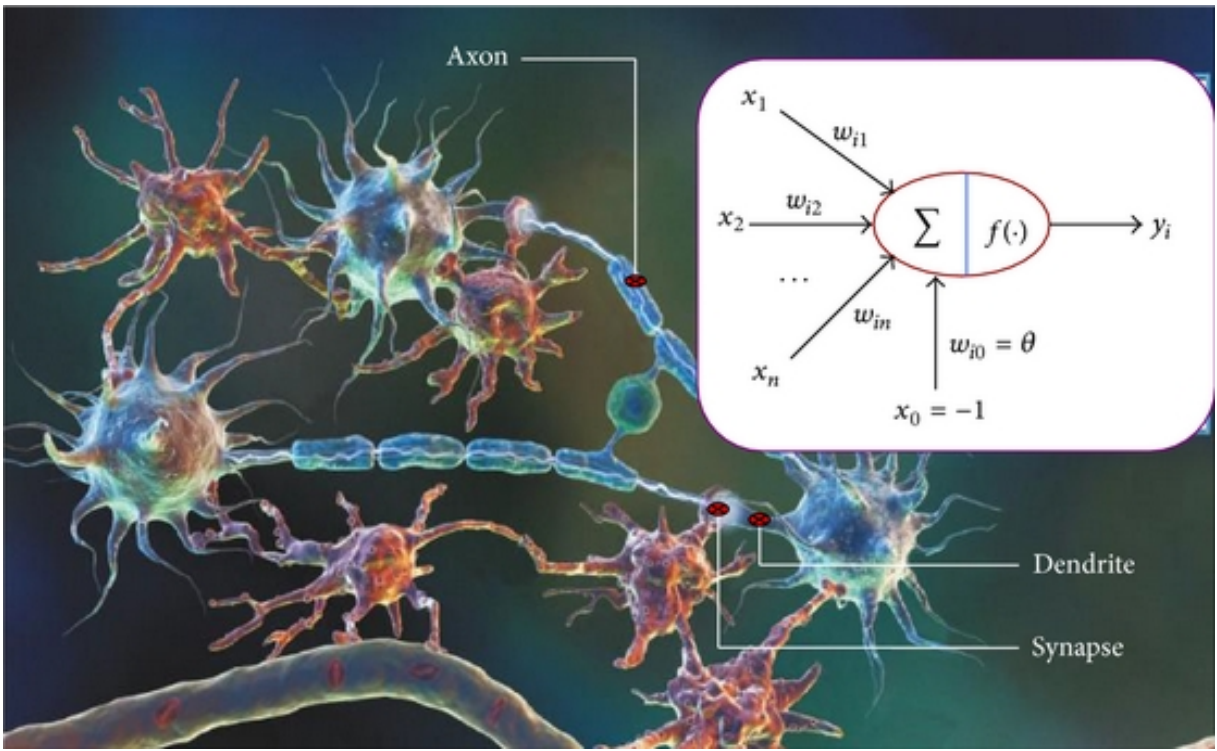

Figure 1. Multiple cells of a BNN architecture with a single unit of an ANN in the inset. Licensed under CC-BY 4.0. Courtesy Wikimedia and Lai, et al. 2015.

The typical representation of ANNs mentioned originates with McCulloch and Pitts (1943), but there have been other types of representation that stress the synaptic and dendritic aspects of BNNs. These include approaches such as dendritic computing (Poirazi and Mel, 2001;

London and Hausser, 2005; Mel, 2008; Jones and Kording, 2021) and sparse distributed representations (Cui et al, 2017). While these alternate forms of ANNs can perform quite capably, they still do not match the more complex behaviors associated with BNNs.

**Heterogeneity and Dynamism of Network Architecture**

In the recent literature on ANNs, the case has been made for introducing BNN-like heterogeneity into otherwise homogeneous ANNs. This case has largely been made with respect to structured architectures (Marblestone, et al. 2016) and anatomical complexity (Richards, et al. 2019). In terms of a structured architecture, ANNs should replicate the multiple systems of function found in a typical BNN. For example, the visual pathways of Mammalian brains involve multiple stages of processes in addition to interactions with attentional, emotional, and autonomic subsystems. These could be made salient in ANNs through the use of interconnected functional modules (Marblestone, et al. 2016). Anatomical complexity is particularly important in terms of enabling plasticity in the service of learning mechanisms. Ideally, learning rules adapted from BNNs may help us avoid the extensive bias and variance sometimes observed in ANNs (Richards, et al. 2019). This may enable durable neural representations that change over the course of learning. The inclusion of stochasticity in ANN function (Florensa, et al. 2017) may also contribute to more BNN-like performance (Harrison, et al. 2005). Taken collectively, this dynamism is a form of temporal heterogeneity, and complements topological heterogeneity by imitating the natural fluctuations of BNNs.

**Organizational Nature of BNNs and ANNs**

Another way to understand the complexity of BNNs (Figure 2) is to conceive of them as complex brain networks (Bullmore and Sporns, 2009). One defining feature network of complex BNN structure is the rich-club organization of connectomes in a number of species. The rich-club connectome has been most comprehensively studied in the nematode *C. elegans* (Towlson, et al. 2013), where a small subset of neurons are connected to most of the neurons in the entire BNN. This results in dense network hubs that exhibit power law behavior (Eguiluz, et al. 2005). One side effect of this type of organization is efficient neuronal processing of high amounts of information at a low energetic cost (Pedersen and Omidvarnia, 2016). In BNNs, the high energy consumption associated with metabolism is offset by hierarchical organization due to a few neurons doing most of the processing (Nigam, 2016). The human brain (a very large BNN) consumes around 20 watts/hour. This would be a good property for ANNs to emulate, as a large ANN running on a Graphics Processing Unit (GPU) consumes between 150 to 250 watts/hour.

The principle of heterogeneity can be realized by flexible network configurations exhibiting highly optimized topologies (Doyle and Carlson, 1999). In BNN, this is achieved using an all-to-all form of connectivity. By contrast, ANN layers are connected in a sequential fashion. In their standard form, this type of connectivity prevents the emergence of a rich-club type of hierarchical information flow. Viewing BNNs as complex networks, where cells are connected in a potentially all-to-all manner, offers a new perspective on ANN configuration. Hinton (2021) has recently proposed the GLOM model, which considers the performance benefits of adding internal structure to ANN topologies. Yet topology is not the only consequence of the complex network approach. Whereas BNNs consist of asynchronous computing nodes, ANN have a more stereotypical structure of computational order and feedback, where the distribution of information flow is uniform across uniform layers of nodes. While traditional backpropagation allows for the improvisation of connection weights, it does not mimic the temporal order of BNNs, which have multiple processes going on at many different timescales.

Another property of BNNs is the role of opponent processes. Opponent processes in BNNs can be defined as physiological or autonomic forms of regulation: as one process shuts down, the other process turns on. This mechanism has been observed in the course of retinal function (Mills, et al. 2014), spatial navigation mechanisms in foraging social insects (Le Moel and Wystrach, 2020), and psychophysiological states such as addiction (Koob and Le Moal, 2008; Radke, et al. 2011). Crucially, this relies upon the existence of processes that are auxiliary to the network structure itself. BNNs as we have defined them are influenced by indirect processes such as metabolism, physiological homeostasis, and even waste disposal mechanisms such as the glymphatic system (Aaling Jessen, et al. 2015) that act to constantly refine a network's structure and function.

**Robust function**
Opponent processes are a theory of function while also being a demonstration of our final comparative category: robustness of function. While the dynamic regulation of complementary processes is necessary for robustness, robust function also enables macro-scale capabilities in a BNN. In BNNs, robustness comes in two forms: fault tolerance and tolerance to ambiguity. Fault tolerance is the ability to overcome errors without compromising function. Regeneration after nervous system damage and invariance to environmental noise exemplify the function of fault tolerance mechanisms in BNNs. This latter example is related to tolerance to ambiguity, which for better or for worse allows BNNs to operate on the basis of intuition and heuristics.

What is the source of this robust function in the network itself? The key is that BNNs are defined by distributed and decentralized network topologies. We can understand this in terms of sensitivity analysis (for BNNs, see Wei, et al. 2018). Through this type of analysis, we find that while no single node is essential for continued network function, some nodes are more important than others. While BNNs are not completely decentralized, functions tend to be distributed across the network. In BNNs, the study of attack tolerance (Albert, et al. 2000; Achard, et al. 2006) shows that small-world network topologies are key to this resilience. While ANNs are also distributed, they do not exhibit the mix of decentralization and hierarchical organization that define BNNs. In BNNs, a decentralised architecture is in part enabled by the ability to self-regenerate (Seguin, et al. 2018), and so may need to be a feature of similarly designed ANNs.

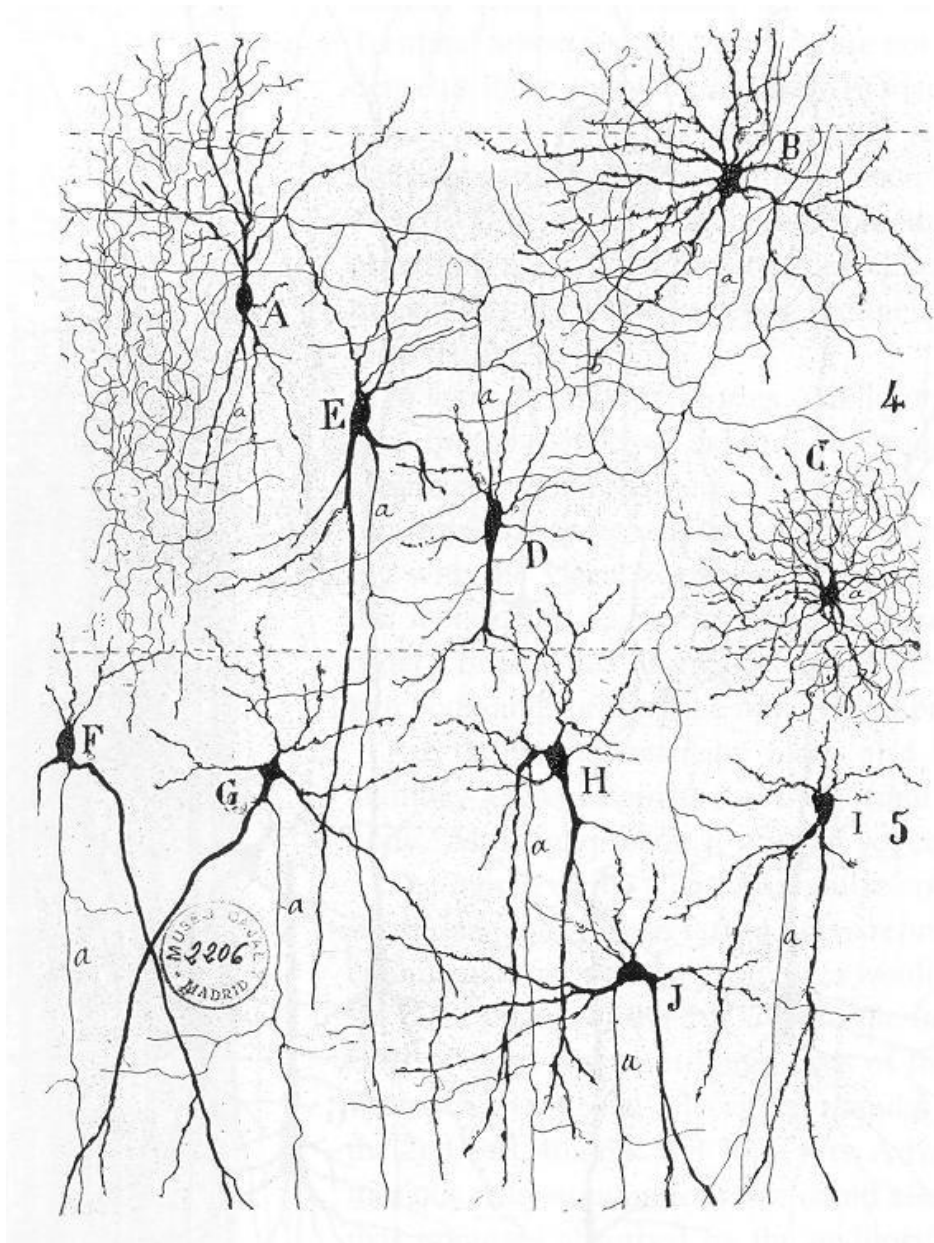

Figure 2. Drawing of many different neuronal morphologies observed in the auditory cortex. From "Texture of the Nervous System of Man and the Vertebrates" by Santiago Ramon y Cajal. Courtesy: Wikimedia.

**Why should we emulate BNNs?**

After evaluating the differences between ANNs and BNNs, one might inquire as to the point of such a comparison. After all, if ANNs yield powerful results, it should stand to reason that a weak analogy is sufficient. In fact, in the field of non-neuronal cognition (Baluska and Levin, 2016; Boisseau, et al. 2016), a similar interpretive analogy exists between functional biological networks on the one hand and intelligent, goal-directed outputs on the other. There are three analytical observations that show the parallels between ANNs and BNNs are more than simple analogies. The first of these is the use of common statistical techniques to analyze both types of networks. Both representational similarity analysis and (to a lesser extent) Bayesian approaches (Barrett, et al. 2019) allows us to understand the representations that are enabled by network topology. According to Petri, et al (2019), ANNs and BNNs must both manage a tradeoff between interactive and independent parallelism. Stated in terms of function, this is a relationship between generalizability of learning and efficient operation. Finally, ANNs can be mathematically reformulated as a directed graph, which enables the type of goal-directedness observed as outputs of a BNN (Guresen and Kayakutlu, 2011).

Another reason to emulate BNNs in the design of ANNs is to take advantage of the information processing capabilities of biological systems. We might ask the following question: is the computational power of connectionist systems due to its analogies with the brain (and cognition) or its analogies with complex systems? If the answer is the latter, then many systems that exhibit intelligent behavior without brains might serve to unify our understanding of the universals of intelligence by drawing parallels between ANNs and more exotic types of computational models such as reservoir computers or liquid state machines.

**ANNs as a phenotype for data**

Perhaps one way to make ANNs more realistic is to create embodied ANNs. Imagine if an ANN was connected to a virtual head, tail, and fins, similar to the axial organization of a BNN, its bilateral inputs, and its posterior outputs (see Figure 3). This provides both the basis for category formation and temporal causality, and is a consequence of the embodied nature of the embedding phenotype. From the standpoint of emulating BNNs, adding in anatomical context this introduces defined systems and pathways where behaviors are regulated over time and recognizable behaviors originate. The current trend in Artificial Intelligence is to create ever-more complex models. In the aforementioned case of GPT-3, models are expansive and are opaque in terms of explainability. While explainability is indeed a field of study, another concern is the lack of plausible behavior. One way to introduce such plausible behavior is to embed these networks in an embodied context. By creating constraints through physical or mechanical interactions with the environment, we may be able to provide enough rigidity in an ANN to steer the system output towards more plausibly realistic behavior.

**Summary**

We have assessed three major differences between ANNs and BNNs: representational complexity, dynamism and organizational structure, and robust function. In doing so, we have provided a number of areas of inquiry for future research. Particularly, we highlight the essential connection between regulatory systems, morphological context, and BNNs throughout the paper.

Do ANNs require a support system to exhibit more lifelike behavioral outputs? To simulate this, we might consider hybrid computational models which combine ANNs with either models of reinforcement learning or evolutionary computation. A framework for some of these models already exists. One example comes from Hasson, et al. (2020), who established a connection between a potential multiobjective evolutionary fitness function in BNNs that is similar to how model parameters of an ANN are optimized. Another way to make ANNs more lifelike is to model ANNs with many of the physiological components of BNNs. This form of biological realism provides us with artificial systems that emulate simple nervous systems (Gleeson, et al. 2018; Sarma, et al. 2018) but not an analysis of large amounts of data.

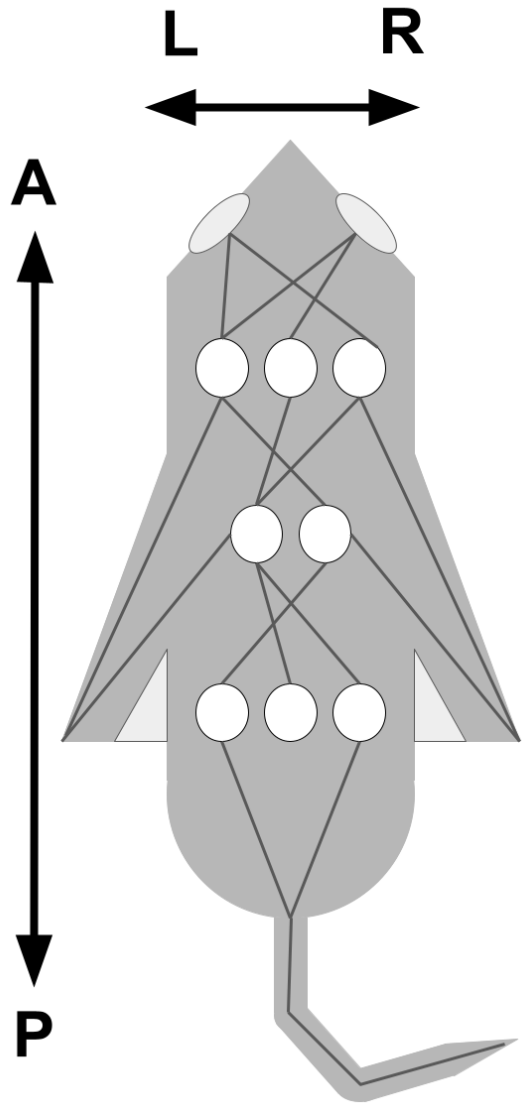

Figure 3. A cartoon diagram of a phenotype for data similar to the developmental phenotypes in Alicea, et al (2021). A network topology is embodied in an animal pseudo-phenotype. In this example, there are two anatomical dimensions that provide spatial context: anterior-posterior

(A-P) and left-right (L-R). The network has direct information (feedback) from the external environment, which might represent integration across different types of data, or different aspects of the same dataset. Input from the environment is bilateral and located in anterior portions of the network, while the output is oriented towards the tail (posterior end of the network). This provides both categorical and causal context, which may be used in lieu of supervision during learning.

Yet perhaps we can approach biological realism by using alternate strategies to implement ANNs rather than modifying the architecture. For example, unsupervised models optimized to imitate broad biological functions can perform similarly to functions such as visual processing and sensory learning (Zhuang, et al. 2021). Another approach called goal-driven deep learning utilizes a different objective than the typical ANN, and yields results more similar to BNNs (Yamins and DiCarlo, 2018). There are many research opportunities for Artificial Life researchers to contribute to understanding how we can close the gap between ANNs and BNNs, in addition to exploiting the details of BNNs to augment current ANN architectures.

**Acknowledgments**. We would like to thank the Neuromatch conference (https://neuromatch.io/) for the opportunity to develop these ideas in the form of a talk. Additional thanks go to the DevoWorm group (https://devoworm.weebly.com/) for feedback and perspective on this topic.